\documentclass[letterpaper]{aamas2017}
\usepackage{algorithmic}
\usepackage{algorithm}
\usepackage{graphicx}
\usepackage{caption}
\usepackage{subcaption}
\usepackage{hyperref}
\usepackage{epstopdf}

\title{Inverse Reinforcement Learning Under Noisy Observations}

\numberofauthors{1}

\author{
\alignauthor Shervin Shahryari and Prashant Doshi\\
\affaddr Institute for AI, University of Georgia, Athens, GA 30602\\
\email pdoshi@cs.uga.edu
}

\begin{document}

\maketitle

\begin{abstract}
We consider the problem of performing inverse reinforcement learning when the trajectory of the expert is not perfectly observed by the learner. Instead, a noisy continuous-time observation of the trajectory is provided to the learner. This problem exhibits wide-ranging applications and the specific application we consider here is the scenario in which the learner seeks to penetrate a perimeter patrolled by a robot. The learner's field of view is limited due to which it cannot observe the patroller's complete trajectory. Instead, we allow the learner to listen to the expert's movement sound, which it can also use to estimate the expert's state and action using an observation model. We treat the expert's state and action as hidden data and present an algorithm based on expectation maximization and maximum entropy principle to solve the non-linear, non-convex problem. Related work considers discrete-time observations and an observation model that does not include actions. In contrast, our technique takes expectations over both state and action of the expert, enabling learning even in the presence of extreme noise and broader applications.
\end{abstract}

%-----------------------------------------------------------------
\section{Introduction}
\label{sec:intro}
%------------------------------------------------------------------

Inverse reinforcement learning (IRL)~\cite{ng2000algorithms} problems seek to find the observed expert's rewards, and usually model the expert as a Markov decision process (MDP)~\cite{puterman2014markov}. Moreover, most methods assume that the learner has perfect observability of the expert's trajectory consisting of a sequence of state and actions~\cite{ng2000algorithms,Abbeel:2004:ALV:1015330.1015430}. In this paper, we relax this assumption -- the learner is not able to observe expert's states and actions directly. Consider the scenario introduced by Bogert and Doshi~\cite{Bogert:2014:MIR:2615731.2615762}, in which an intruder wants to learn a patroller's behavior in order to penetrate the patrol without being spotted. In order to do so, the intruder (learner) must be hidden from the patroller's view. Therefore, it would not be able to see the patroller directly most of the time. Instead, it may hear its movement sound at all times. Consequentially, the learner can estimate the patroller's state and action using an observation model.

\begin{figure}[!t]
\centering
\includegraphics[width=1.0\linewidth]{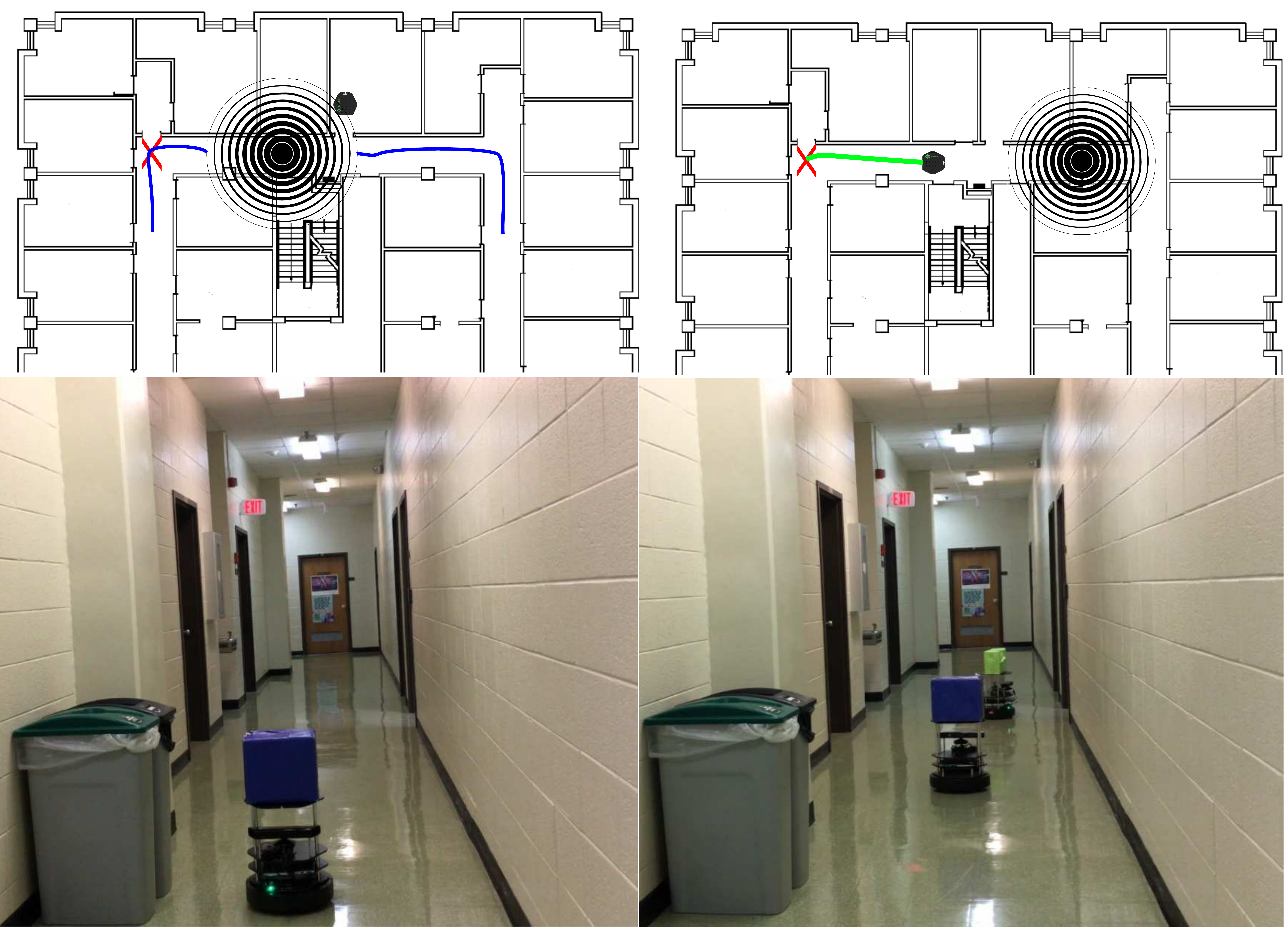}
\caption{\small The top two images are our experiment in the simulation. We used ROS to run the simulation. The patroller securing the hallway and the concentric circles around the patroller is an indication of the magnitude of the sound intensity. As the distance between the patroller and the intruder increases, the intensity (thickness of the circle boundary) decreases. The lower images are from the physical experiment done with two Turtlebots. The two images on the right-hand side show the moment that the intruder starts moving, and the two images on the left-hand side show the observated expert as it patrols.}
\label{fig:col}
\end{figure}

In the above example, the complete trajectory of state-action pairs is not seen by the learner. However, a sequence of observations is provided to the learner. In this case, the observation is the  intensity over time of the movement sound of the expert manifesting as a robot. The sound intensity is inversely proportional to the squared distance from the sound source~\cite{lighthill1952sound}. This makes it possible for the learner to infer the state and action of the expert on receiving a sequence of sound intensities. However, due to surrounding noise, this observation could be noisy.

Specifically, we consider the problem of learning the behavior of a robot that patrols a known perimeter using maximum entropy IRL. We assume that the environment state is fully observable to the expert, hence the expert is aware of its own action and state. However, the state and action of the expert are not observed by the learner; instead, a noisy observation of the state and action is provided to the learner.

This paper makes the following contribution:
\begin{enumerate}
  \item We generalize IRL to operate under situations in which the observation of the expert received by the learner has considerable amount of noise.
  \item We incorporate an observation model into IRL, which considers time-extended observations as a function of both state and action. We generalize expectation-maximization for IRL~\cite{wang2012latent,
  	Bogert:2016:EIR:2936924.2937076}, which allows the trajectory of the expert to be hidden from the learner, with this observation model. This generalization enables the learner to fuse data from different sensors with different levels of noise.
\end{enumerate}

Rest of this paper is organized as follows. We review IRL and the maximum entropy method in Sec.~\ref{sec:bac}. Observation model and a generalization of maximum entropy IRL (Robust-IRL) in the context of noisy observation are presented in Sec.~\ref{sec:RIRL}. We provide a detailed algorithm of robust-IRL in Sec.~\ref{sec:alg}. Finally, we evaluate the performance of the Robust-IRL on two robotic problem domains in Sec.~\ref{sec:eval}. After that, we discuss related work in Sec.~\ref{sec:rw}. In Sec.~\ref{sec:con} we conclude this paper.

\section{Background}
\label{sec:bac}

\subsection{Inverse Reinforcement Learning}
\begin{figure}[t]
\centering
\includegraphics[width=0.8\linewidth]{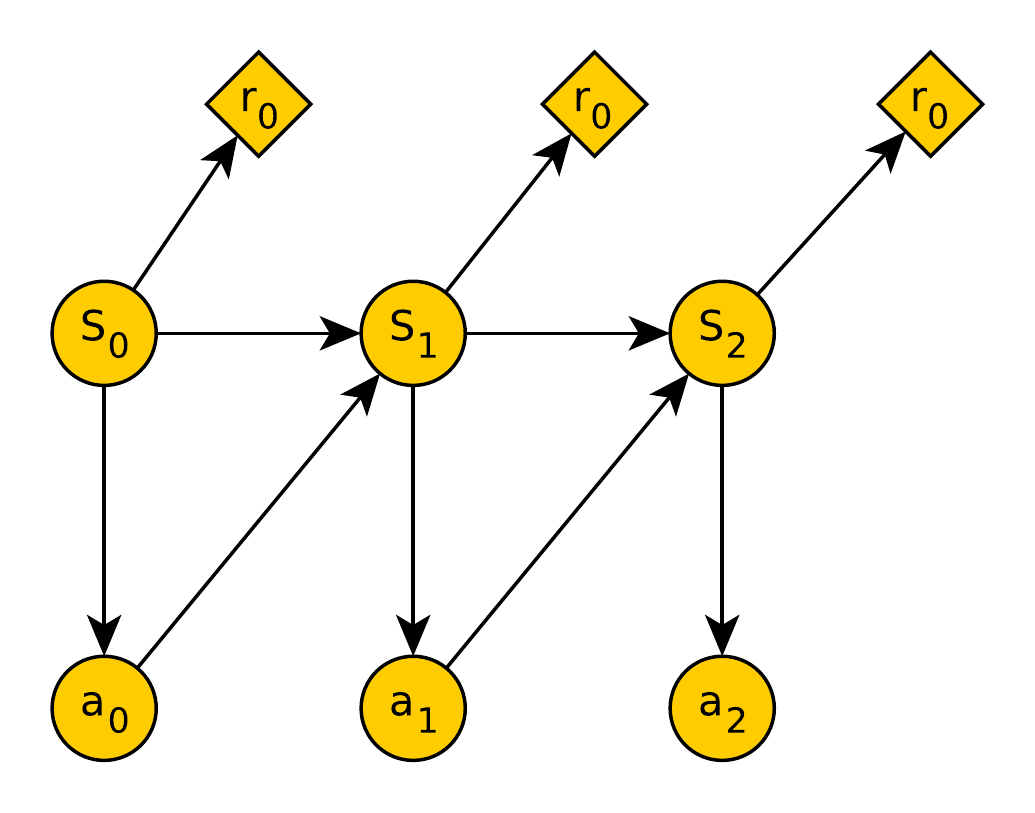}
\caption{An illustration of the expert's MDP considering 3 horizons. $s_i$ , $a_i$ and $r_i$ shows the state, action and reward of the agent at time step $i$.}
\label{fig:MDP}
\end{figure}

IRL seeks to find the most likely reward function $R_E$, which an expert $E$ is executing~\cite{gao2012survey}~\cite{Abbeel:2004:ALV:1015330.1015430}~\cite{ng2000algorithms}~\cite{neu2012apprenticeship}. Current IRL methods assume the presence of a single expert that solves a Markov decision process (MDP). Moreover, they assume that the MDP is fully known and observable by the learner except for the reward function. Since the state and action of the expert is fully observable for the learner, it can construct a trajectory of an arbitrary length, which is consist of a sequence of state and action pairs,$T =(<s,a>^0 , <s,a>^1 , ... , <s,a>^L)$ , where  $s$ and $a$ belong to a set of possible states and action of the expert respectively and $T$ belongs to a set of observed trajectories $\tau$, and $\mathcal{T}$ is the finite set of all trajectories of length L; $\tau \subseteq \mathcal{T}$. Figure~\ref{fig:MDP} is an illustration of a typical MDP for an expert.

Since the space of the possible reward function is outsized, it is common to represent the reward function as a linear combination of $K > 0$ binary features. $R_{E}(s,a) = \Sigma^{K}_{1}\theta_{k}\phi_{k}(s,a)$. $\theta_{k}$ are weights and $\phi_{k}(s,a) \rightarrow \left\{0,1\right\}$ is a binary feature function that maps a pair of state and action to either zero (not activated) or 1 (activated)~\cite{Abbeel:2004:ALV:1015330.1015430}. The reward function $R_{E}(s,a)$ takes in a pairs of state and action and maps it to a real value number by using the feature function. Current state of the art IRL algorithms utilize feature expectation as a way of evaluating the quality of learned reward function~\cite{Abbeel:2004:ALV:1015330.1015430}. One can use the following formulation to calculate the $k^{th}$ feature expectation for a learned reward function $R_{E}$: $\sum\limits^{}_{T \in \mathcal{T}}Pr(T) \sum\limits^{}_{<s,a> \in T}\phi_{k}(s,a)$. The exceptions are compared with those which are calculated based on the expert trajectory. The feature exception of the expert's trajectory can be calculated as follows: $\hat{\phi}_{k} = \sum\limits^{}_{T \in \theta}\tilde{Pr}(T)\sum\limits^{}_{<s,a> \in T}\phi_{k}(s,a)$, where $\tilde{Pr}(T)$ is the empirical probability of trajectory $T$.

\subsection{Maximum Entropy IRL}
Inverse reinforcement learning is an ill-posed problem~\cite{ng2000algorithms}; it means that there is more than one reward function that can explain the expert's behavior.  State of the art algorithm max-Ent IRL, proposed by Ziebart \textit{et al.}~\cite{ziebart2008maximum}, utilizes maximum entropy principle to mitigate this ill-posed problem.

Max-Ent IRL recovers a distribution over trajectories such that it has the maximum entropy among all distributions of trajectories that match the observed feature expectation. Mathematically this problem can be formulated as a convex nonlinear optimization~\cite{ziebart2008maximum}:
\begin{equation} \label{IRL-program}
\begin{aligned} 
&\max_{\Delta}(-\sum\limits^{}_{T \in \mathcal{T}}Pr(T)\log Pr(T))\\
&subjected \ to \\
&\sum\limits^{}_{T \in \mathcal{T}}Pr(T) = 1 \\
&\sum\limits^{}_{T \in \mathcal{T}}Pr(T)\sum\limits^{}_{<s,a> \in T}\phi_{k}(s,a) = \hat{\phi_{k}}
\end{aligned}
\end{equation}
where $\Delta$ is the space of all possible $Pr(T)$. In order to solve this optimization problem we can apply Lagrangian relaxation to bring both constrains into the objective function and then solve the dual utilizing exponentiated gradient descent.
\begin{equation} \label{IRL-Lagrangian}
\begin{aligned}
\mathcal{L}(Pr(T),\theta,\eta) &= -\sum\limits^{}_{T \in \mathcal{T}}Pr(T)\log Pr(T) + \sum\limits^{}_{k}\theta_{k} \\
&(\sum\limits^{}_{T \in \mathcal{T}}Pr(T)\sum\limits^{}_{<s,a> \in T}\phi_{k}(s,a) - \hat{\phi}_{k})\\
& + \eta (\sum\limits^{}_{T \in \mathcal{T}}Pr(T) - 1)
\end{aligned} 
\end{equation}
Now we take the partial derivative  with respect to $Pr(T)$ and set it to zero to find the optimal value:
\begin{equation} \label{IRL-Lagrangian partial derivative}
\begin{aligned}
\frac{\partial \mathcal{L}}{\partial Pr(T)} = -\log Pr(T) - 1 + \sum\limits^{}_{k}\theta_{k}\sum\limits^{}_{<s,a> \in T}\phi_{k}(s,a)+\eta = 0
\end{aligned} 
\end{equation}
Solving Eq.~\ref{IRL-Lagrangian partial derivative} for $Pr(T)$ we have:
\begin{equation} \label{IRL-Pr(T)}
\begin{aligned}
Pr(T) = \frac{e^{\sum\limits^{}_{k}\theta_{k}\sum\limits^{}_{<s,a> \in T}\phi_{k}(s,a)}}{n(\theta)}
\end{aligned} 
\end{equation}
where $n(\theta)$ is the normalizing factor. By plugging Eq.~\ref{IRL-Pr(T)} into Eq.~\ref{IRL-Lagrangian} we get Eq.~\ref{IRL-dual}:
\begin{equation} \label{IRL-dual}
\begin{aligned}
\mathcal{L}^{dual}(\theta) = \log(n(\theta)) - \sum\limits^{}_{k}\theta_{k}\hat{\phi}_{k}
\end{aligned} 
\end{equation}
Eq.~\ref{IRL-dual} is the dual program, which can be solved by using the exponentiated gradient descent to find the optimal values of $\theta$. Eq.~\ref{IRL-dual gradient} is the gradient.
\begin{equation} \label{IRL-dual gradient}
\begin{aligned}
\nabla \mathcal{L}^{dual}(\theta) = \sum\limits^{}_{T \in \mathcal{T}}Pr(T)\sum\limits^{}_{<s,a> \in T}\phi_{k}(s,a) - \hat{\phi}_{k}
\end{aligned} 
\end{equation}
As it shown above, calculating the gradient involves summing over the set of all possible trajectories that may be intractable in most of the problems. However Ziebart \textit{et al.}~\cite{ziebart2008maximum} proposed an efficient approach that calculates the expected edge frequency (state visitation frequency). 

%----------------------------------------------------------------------------------
\section{Robust inverse reinforcement learning}
\label{sec:RIRL}
%----------------------------------------------------------------------------------

In this paper, we consider a situation when the sensory information is noisy or information comes from different sensors with different levels of noise. A simple modification of the maximum entropy inverse reinforcement learning is to utilize an observation model and construct the most probable trajectory. As a result, this method would not be appropriate when the amount of noise is considerable, therefore it motivates a principled way of handling the noise in observation.

On the other hand, we can maintain a distribution over possible trajectories given the observation. We propose a principled way to handle the noise through observation model and expectation maximization. Then we apply our method to two specific cases: first when the observation comes from one noisy sensor (microphone), second when the observation comes from two sensors with two different levels of noise (microphone and camera).

\subsection{Hidden MDP}
We consider a setting, where the learner receives the sound intensity of the expert's movement sound as observations instead of observing expert's state and action directly. This motivated by the application of utilizing noisy sensory data.
    
\begin{figure}
\centering
\includegraphics[width=1\linewidth]{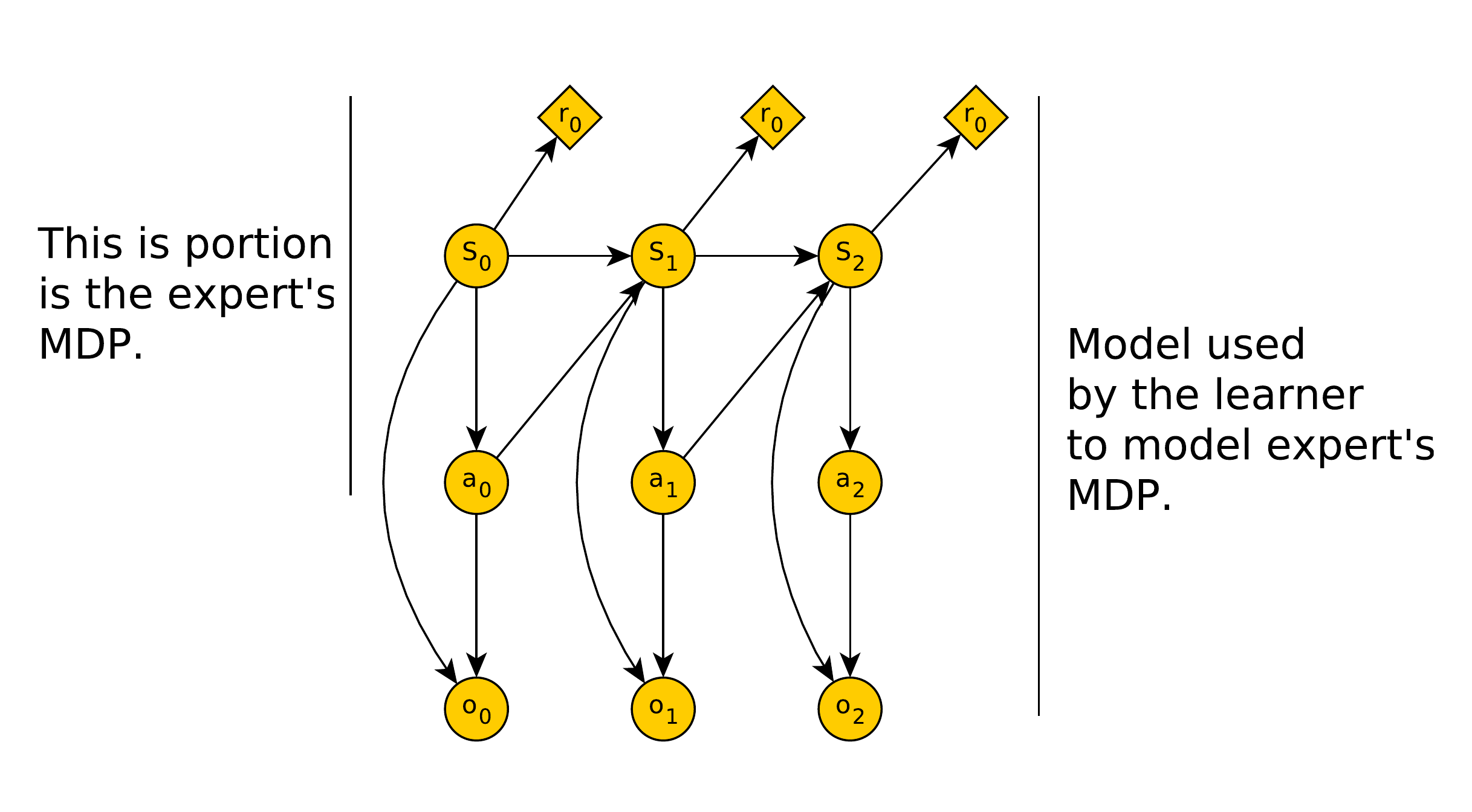}
\caption{\small In hMDP, the state and action are hidden from the learner but an observation of state and action is provided at each time step}
\label{fig:hMDP}
\end{figure}

As we mentioned above the expert solves a Markov decision process to construct its policy. However, since the learner cannot observe expert's state and action it cannot model the expert as an MDP. We adopt the hMDP framework, proposed by Kitani \textit{et al.}~\cite{kitani2012activity}, to model the expert from learner's perspective. Unlike the hMDP proposed by Kitani \textit{et al.} our adoption of hMDP incorporates actions into the observation model. Figure~\ref{fig:hMDP} illustrates our proposed hMDP.

\subsection{Observation Model} \label{sec:obs}
As one may notice observation model plays a crucial role in robust inverse reinforcement learning. Incorporating actions into the observation model introduces challenges, which rises from considering observation in discrete time. Using sound intensity makes it easy to infer the state, however in order to infer the action we must have two consecutive sound intensities. For instance, if the intensity decreases we can infer that the expert most probably moved towards the learner or if the opposite is true the expert moved away from the learner. In other words, for inferring the current action we must have the previous observation (previous sound intensity) in addition to the current observation (current sound intensity), therefore, the Markovian assumption would not hold anymore. This problem arises from the fact that observations are considered in discrete time steps. We can handle this issue in the following manner. 

In our case, we use sound intensities as observations. Sound intensity is inversely proportional to the square distance of the sound source to the listener, $I = \frac{k}{R^{2}}$~\cite{lighthill1952sound}. In the proposed hMDP, state and action happen in discrete time steps (decision epochs), however, observation evolves in continuous time within each decision epoch. Since the observation evolves in continuous time, we can represent observation at each decision epoch by a function $f(t)$, where $t$ represents continuous time. 
\newtheorem{theorem}{Theorem}
\begin{theorem}\label{th-f(t)}
Let $k, \ a, \ b \ and \ c$ be constant then the structure of $f(t)$ is as follows:
\begin{equation} \label{RIRL-f(t)}
\begin{gathered} 
f(t) = \frac{k}{at^{2}+bt+c}
\end{gathered}
\end{equation}
\end{theorem}

Modeling observation in continuous time has advantages over other approaches. Imagine a scenario in which the time of the observation is stochastic; in this scenario, the length of the decision epoch serves as bound for the observation time. The learner receives some samples in a decision epoch and uses regression to find the function that explains observation over continuous time. Consequentially, this approach recovers information about the time that there is no sample provided for the learner.

\begin{figure}
\centering
\includegraphics[width=1\linewidth]{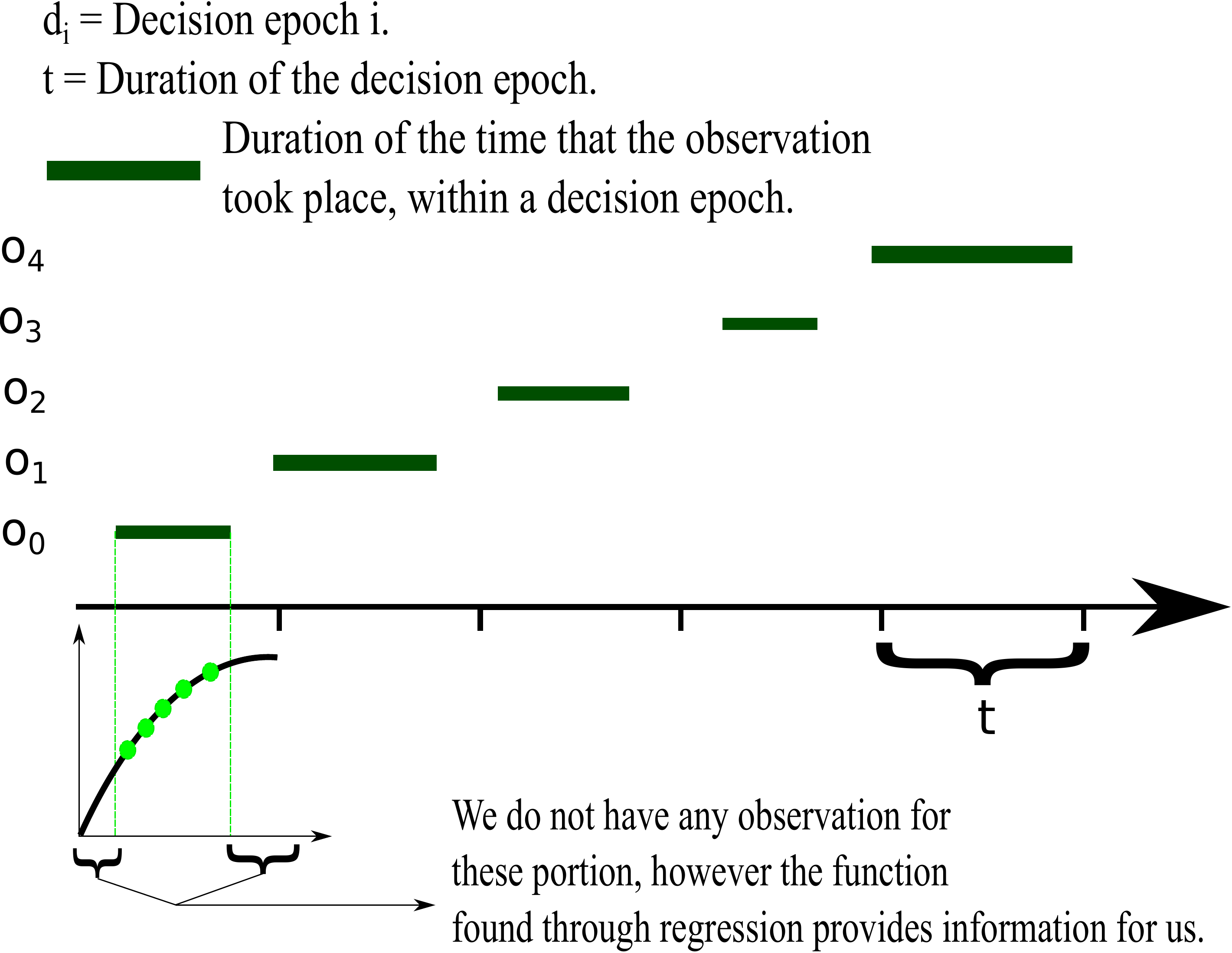}
\caption{\small Illustration of f(t). $d_i$ shows the end of each decision epoch $i$. $t$ is the continues time and each green bar shows the portion of the decision epoch which the sound intensity samples are provided.}
\label{fig:f(t)}
\end{figure}

Figure~\ref{fig:f(t)} illustrates how we can recover information by extrapolating or interpolating when there are not many samples in that decision epoch.

\subsection{Formulation}
In inverse reinforcement learning, the observed trajectory of length $L$ is $T =(<s,a>^0 , <s,a>^1 , ... , <s,a>^L)$. As we mentioned before, in our case, this trajectory is not provided for the learner. Instead a sequence of observation of length $M$, $\vec{\omega} = (o^{0} , o^{1} , ... , o^{M})$, is provided. Lets consider the sequence of observations $\vec{\omega}$ as $Y$, the observed data, and the trajectory $T$ as $Z$, hidden data. In other words $X = (Y \cup Z)$, where the $X$ is the total data.

Simply one can utilizes observation model $Pr(o^{i}|<s,a>)$ to calculate the most likely state and action pair at the time step $i$ with out considering time step $i-1$. However, this approach totally disregards the effect of the transition function and the policy of the expert in constructing the trajectory. In contrast, we propose a revised formulation of maximum entropy inverse reinforcement learning that allows an expectation over trajectories (hidden data) given the sequence of observations. This method allows considering the effect of the transition function and expert's policy in constructing a distribution over possible trajectories.
\begin{equation} \label{RIRL-program}
\begin{aligned} 
&\max_{\Delta}(-\sum\limits_{\vec{\omega},T}Pr(\vec{\omega},T)log(Pr(\vec{\omega},T))\\
&subjected \ to\\
&\sum\limits_{\vec{\omega},T}Pr(\vec{\omega},T)=1\\
&\sum\limits_{\vec{\omega} \in \Omega}\sum\limits_{T \in \tau}Pr(\vec{\omega},T) \sum\limits_{(s,a) \in T} \phi_{k}(s,a) = \hat{\phi_{k}}
\end{aligned}
\end{equation}
where:
\begin{equation} \label{RIRL-Pr(w,T)}
\begin{gathered} 
Pr(\vec{\omega},T) = Pr(\vec{\omega}|T)Pr(T)
\end{gathered}
\end{equation}
\begin{equation} \label{RIRL-Pr(T)}
\begin{gathered} 
Pr(T)=Pr(s_{0}) \prod^{n-1}_{i=1}Pr(s_{i+1}|s_{i},a_{i})Pr(a_{i}|s_{i})
\end{gathered}
\end{equation}
\begin{equation} \label{RIRL-feature exp}
\begin{gathered} 
\hat{\phi_{k}} = \frac{1}{|\tilde{\Omega|}}\sum\limits_{\vec{\omega} \in \tilde{\Omega}} \sum\limits_{T}Pr(T|\vec{\omega})\sum\limits_{(s,a) \in T} \phi_{k}(s,a)
\end{gathered}
\end{equation}
\begin{equation} \label{RIRL-Pr(T|w)}
\begin{gathered} 
Pr(T|\vec{\omega})=\eta Pr(\vec{\omega}|T)Pr(T)
\end{gathered}
\end{equation}

Where, $\Delta$ is the space of all distribution $Pr(\vec{\omega},T)$ and $\Omega$ is the set of sequence of observations from learner's view. In other words $\vec{\omega} \in \Omega$.

Given the above we can apply Lagrangian relaxation to bring both constraints into the objective function, however, because of the presence of conditional probability in the Lagrangian $\mathcal{L}(Pr(\vec{\omega},T);\theta;\eta)$ , the relaxed objective function is not convex.

\begin{equation} \label{RIRL-Lagrangian partial derivative exact}
\begin{aligned}
&\frac{\partial \mathcal{L}(Pr(X),\theta)}{\partial  Pr(X)} = -log(Pr(X))-1&\\
&+ \sum\limits_{k} \theta_{k} \sum\limits_{(s,a)} \phi_{i}(s,a) + \sum\limits^{K}_{k=1}\theta_{k} \bigg(\sum\limits^{}_{Y \in \Omega}\tilde{Pr}(Y)&\\
& \frac{\sum\limits^{}_{Z^{\prime} \in \mathcal{Z}}\bigg[\ \sum\limits^{}_{<s,a> \in X^{\prime}} \phi_{k}(s,a) - \sum\limits^{}_{<s,a> \in X}\phi_{k}(s,a) \bigg]Pr(X^{\prime})}{Pr(Y)^{2}})\bigg)&\\
&+ \eta
\end{aligned}
\end{equation}
where $Y = T$, $Z = \vec{\omega}$, $X = (Y \cup Z)$ and $X^{\prime} = (Y \cup Z^{\prime})$. As one can see, there is no closed form solution for Eq.~\ref{RIRL-Lagrangian partial derivative exact}. However, Wang \textit{et al.}~\cite{wang2012latent} proposed an approximation with the following form.
\begin{equation} \label{RIRL-Lagrangian partial derivative wang}
\begin{aligned}
\frac{\partial \mathcal{L}(Pr(\vec{\omega},T),\theta)}{\partial  Pr(\vec{\omega},T)} \approx -log(Pr(\vec{\omega},T))-1 + \sum\limits_{k} \theta_{k} \sum\limits_{(s,a)} \phi_{i}(s,a) + \eta
\end{aligned}
\end{equation}

Setting Eq.~\ref{RIRL-Lagrangian partial derivative wang} to zero, we have:
\begin{equation} \label{RIRL-pr(w,T)}
\begin{aligned}
Pr((\vec{\omega},T),\theta) \approx \frac{e^{\sum\limits_{k} \theta \sum\limits_{(s,a)} \phi_{i}(s,a)}}{n(\theta)}
\end{aligned}
\end{equation}
where $n(\theta)$ is the normalizer constant.

Now we can approximately calculate the optimal value of $Pr(\vec{\omega},T)$, therefore we can calculate the value of Lagrangian parameters. By plugging the above equation to $\mathcal{L}(Pr(\vec{\omega},T);\theta;\eta)$ we will have:
\begin{equation} \label{RIRL-dual}
\begin{aligned}
\mathcal{L}^{(Dual)}(\theta) = log(n(\theta)) - \sum\limits_{k} \theta_{k} \sum\limits_{\vec{\omega} \in \tilde{\Omega}} \frac{1}{|\tilde{\Omega}|} \sum\limits_{T} Pr(T|\vec{\omega}) \sum\limits_{(s,a)} \phi(s,a)
\end{aligned}
\end{equation}

\subsection{Expectation-Maximization}
Because of the presence of $Pr(T|\vec{\omega})$ in the Eq.~\ref{RIRL-dual}, we cannot use exponentiated gradient descent to obtain the optimal value of parameter vector. However Wang \textit{et al.}~\cite{wang2012latent} proposed an iterative EM approach that can be adapted to our Maximum entropy model to find the optimum value of vector parameter.

Likelihood of Lagrangian parameter can be defined as follows:
\begin{equation} \label{RIRL-ll}
\begin{aligned}
LL(\theta|\vec{\omega}) &= \log \prod^{}_{\vec{\omega} \in \Omega}Pr(\vec{\omega};\theta)^{\tilde{Pr}(\vec{\omega})}\\
& = \sum\limits^{}_{\vec{\omega} \in \Omega}\tilde{Pr}(\vec{\omega})\log Pr(\vec{\omega};\theta) \sum\limits^{}_{T \in \mathcal{T}}Pr(T|\vec{\omega};\theta)\\
& = \sum\limits^{}_{\vec{\omega} \in \Omega}\tilde{Pr}(\vec{\omega}) \sum\limits^{}_{T \in \mathcal{T}}Pr(T|\vec{\omega};\theta)\log Pr(\vec{\omega};\theta)
\end{aligned}
\end{equation}

Rewriting $ Pr(\vec{\omega}|\theta)$ as $\frac{Pr(\vec{\omega},T;\theta)}{Pr(T|\vec{\omega},\theta)}$ in Eq.~\ref{RIRL-ll}:

\begin{equation} \label{RIRL-ll for Em}
\begin{aligned}
LL(\theta|\vec{\omega}) &= \sum\limits^{}_{\vec{\omega} \in \Omega}\tilde{Pr}(\vec{\omega}) \sum\limits^{}_{T \in \mathcal{T}}Pr(T|\vec{\omega};\theta)\log \frac{Pr(\vec{\omega},T;\theta)}{Pr(T|\vec{\omega},\theta)}\\
&= \sum\limits^{}_{\vec{\omega} \in \Omega}\tilde{Pr}(\vec{\omega}) \sum\limits^{}_{T \in \mathcal{T}}Pr(T|\vec{\omega};\theta)(\log (Pr(\vec{\omega},T;\theta))\\
& - \log(Pr(T|\vec{\omega},\theta)))
\end{aligned}
\end{equation}
Now we may use EM to improve the likelihood in Eq.~\ref{RIRL-ll for Em} iteratively. We can reformulate the likelihood as $Q(\theta,\theta^{t}) + C(\theta,\theta^{t})$ where:
\begin{equation} \label{RIRL-Q}
\begin{aligned}
Q(\theta,\theta^{t}) = \sum\limits^{}_{\vec{\omega} \in \Omega}\tilde{Pr}(\vec{\omega}) \sum\limits^{}_{T \in \mathcal{T}}Pr(T|\vec{\omega};\theta^{t})\log (Pr(\vec{\omega},T;\theta))
\end{aligned}
\end{equation}
\begin{equation} \label{RIRL-C}
\begin{aligned}
C(\theta,\theta^{t}) = -\sum\limits^{}_{\vec{\omega} \in \Omega}\tilde{Pr}(\vec{\omega}) \sum\limits^{}_{T \in \mathcal{T}}Pr(T|\vec{\omega};\theta^{t})\log(Pr(T|\vec{\omega},\theta))
\end{aligned}
\end{equation}

Replacing $Pr(\vec{\omega},T;\theta)$ in Eq.~\ref{RIRL-Q} with Eq.\ref{RIRL-Pr(w,T)}:
\begin{equation} \label{RIRL-Q dual negative}
\begin{aligned}
Q(\theta,\theta^{t}) & = -(log(n(\theta)) - \\
& \sum\limits_{k} \theta_{k} \sum\limits_{\vec{\omega} \in \tilde{\Omega}} \frac{1}{|\tilde{\Omega}|} \sum\limits_{T} Pr(T|\vec{\omega}) \sum\limits_{(s,a)} \phi(s,a))
\end{aligned}
\end{equation}

One may notice that $Q$ function is the negative of the dual presented in Eq.~\ref{RIRL-dual}. Therefore maximizing the $Q$ function is equivalent to minimizing the dual. Using these facts, we may reformulate the original problem stated in Eq.~\ref{RIRL-program} as follows.

\subsubsection{E-step:}
In the E-step we use the parameter $\theta^{t}$ from the previous iteration to calculate the feature expectation of the expert.
\begin{equation} \label{RIRL-E-step}
\begin{aligned}
\hat{\phi}^{T|\vec{\omega},t}_{k} = \sum\limits^{}_{\vec{\omega} \in \Omega}\tilde{Pr}(\vec{\omega}) \sum\limits^{}_{T \in \mathcal{T}}Pr(T|\vec{\omega};\theta^{t})\sum\limits^{}_{<s,a> \in T}\phi_{k}(s,a)
\end{aligned}
\end{equation}
To calculate $Pr(T|\vec{\omega})$ we may use Bayes rule.
\begin{equation} \label{RIRL-E-step Bayes rule}
\begin{aligned}
Pr(T|\vec{\omega})=\eta Pr(\vec{\omega}|T)Pr(T)
\end{aligned}
\end{equation}
where:
\begin{equation} \label{RIRL-E-step Pr(T)}
\begin{aligned}
Pr(T)=Pr(s_{0}) \prod^{n-1}_{i=1}Pr(s_{i+1}|s_{i},a_{i})Pr(a_{i}|s_{i})
\end{aligned}
\end{equation}
\begin{equation} \label{RIRL-E-step Pr(w|T)}
\begin{aligned}
Pr(\vec{\omega}|T) = \prod^{n}_{i=1}Pr(o_i|s_i,a_i)
\end{aligned}
\end{equation}

In Eq.~\ref{RIRL-E-step Pr(T)}, $Pr(a_i|s_i)$ is the expert's policy give $\theta^{t}$ and in Eq.~\ref{RIRL-E-step Pr(w|T)}, $Pr(o_i|s_i,a_i)$ is the observation model.
\subsubsection{M-step:}
In the M-Step we utilize the feature expectation that has been calculated in the E-Step to obtain the $\theta$.
\begin{equation} \label{RIRL-M-step}
\begin{aligned}
&\max_{\Delta}(-\sum\limits_{\vec{\omega},T}Pr(\vec{\omega},T)log(Pr(\vec{\omega},T))\\
&subjected \ to\\
&\sum\limits_{\vec{\omega},T}Pr(\vec{\omega},T)=1\\
&\sum\limits_{\vec{\omega} \in \Omega}\sum\limits_{T \in \tau}Pr(\vec{\omega},T) \sum\limits_{(s,a) \in T} \phi_{k}(s,a) = \hat{\phi}^{T|\vec{\omega},t}_{k}
\end{aligned}
\end{equation}

As it shown in Eq.~\ref{RIRL-E-step} calculating E-Step involves a summation over all possible trajectories. Calculating this summation is infeasible in real domain problem. We utilize Gibbs sampling~\cite{hastings1970monte} to approximate this summation.

\section{Algorithm}\label{sec:alg}

Following is the complete algorithm of Robust-IRL. Algorithm~\ref{RIRL-al} shows the overall steps in Robust-IRL. Algorithm~\ref{RIRL-al E-step} and \ref{RIRL-al Estep Gibbs Sampling} show a detailed description of the E-step and the Gibbs sampling needed for the E-step respectively.

In algorithm~\ref{RIRL-al}, at line 1 we initialize the reward function randomly and then we construct the optimal policy accordingly at line 2. Then we do E-step (line 5) and M-Step (lines 7-10) repeatedly till convergence.

Algorithm~\ref{RIRL-al E-step} shows the exact solution for the E-step. At lines 3 and 4, we calculate the probability of each trajectory given the transition function, observation model, and current policy. Then, at line 5 we multiply this probability by the feature count of the trajectory and accumulate it into a variable to calculate the feature expectation of the distribution over trajectories that is under consideration. As we mentioned before this might become infeasible in domains with large state and action spaces. Algorithm~\ref{RIRL-al Estep Gibbs Sampling} shows how to approximate the feature expectation using Gibbs sampling.

\begin{algorithm} 
\caption{Robust inverse reinforcement learning}
\begin{algorithmic}[1]
\STATE $RewardWeights\gets$ Initialize
\STATE $Policy\gets$ Initialize
 \WHILE{FeatureException not converged}
  \STATE \textbf{E-step}: 
  \STATE $Feature Exception \gets$ $\sum\limits_{\vec{\omega} \in \tilde{\Omega}} \frac{1}{|\tilde{\Omega}|} \sum\limits_{T} Pr(T|\vec{\omega}, \theta^{(t)}) \sum\limits_{(s,a)} \phi(s,a)$
  
  \STATE \textbf{M-step}:
  \STATE $RewardWeights \gets$ $log(n(\theta)) - \sum\limits_{k} \theta_{k} \sum\limits_{\vec{\omega} \in \tilde{\Omega}} \frac{1}{|\tilde{\Omega}|} \sum\limits_{T} Pr(T|\vec{\omega}) \sum\limits_{(s,a)} \phi(s,a)$
  \STATE
  \STATE update Reward Function
  \STATE update Policy
\ENDWHILE
\end{algorithmic}\label{RIRL-al}
\end{algorithm}

\begin{algorithm}
\caption{E-step}
\begin{algorithmic}[1]
\STATE $Feature Exception \gets$ Initialize to all zero
\FORALL{$T \in \tau$}
    \STATE $Pr(T)=Pr(s_{0}) \prod^{n-1}_{i=1}Pr(s_{i+1}|s_{i},a_{i})Pr(a_{i}|s_{i})$
    \STATE $Pr(T|\vec{\omega})=\eta Pr(\vec{\omega}|T)Pr(T)$
    \STATE $FeatureException = FeatureException + \sum\limits_{\vec{\omega} \in \tilde{\Omega}} \frac{1}{|\tilde{\Omega}|} Pr(T|\vec{\omega}, \theta^{(t)}) \sum\limits_{(s,a)} \phi(s,a)$
  \ENDFOR
\end{algorithmic}\label{RIRL-al E-step}
\end{algorithm}

\begin{algorithm}
\caption{E-step Gibbs Sampling}
\begin{algorithmic}[1]
\STATE $T \gets$ Initialize using Observation and current Policy
\STATE $FeatureExpectationVector \gets$ Initialize to all zero
\WHILE{$FeatureExpectationVector$ not converged} 
	\FORALL{number of sampling steps}
		\STATE Sample one Node in T at random according to its Markov blanket using observations, transition function, observation model and current policy 
		\STATE Update $FeatureExpectationVector$
	\ENDFOR
\ENDWHILE
\end{algorithmic}\label{RIRL-al Estep Gibbs Sampling}
\end{algorithm}

\section{Performance Evaluation:}\label{sec:eval}
In this section, we describe two domains that we used to evaluate the robust-IRL method that described earlier.

\subsection{Metrics and Baseline}
First, we need to discuss the baselines and the method of comparison between Robust-IRL and baseline. At first glance, it is tempting to directly compare the learned reward function with the true reward function. However, this may not be a good metric for comparison because it is possible for two drastically different reward functions to results into a very similar policies. Choi and Kim~\cite{choi2011inverse} proposed to compare behaviors instead of comparing reward functions. For doing so we need to calculate the value function by solving the expert's MDP with the true reward function, then do the same process using the learned reward function. Now the difference between these two value functions indicates the deviation from desired behavior. This metric is called inverse learning error (ILE)~\cite{choi2011inverse}.

\begin{equation} \label{ILE}
\begin{aligned} 
ILE = ||V^{\pi^{L}} - V^{\pi^{E}}||
\end{aligned}
\end{equation}

Where $V^{\pi^{L}}$ is the value function calculated by utilizing policy $\pi^{L}$ on the expert's MDP and $V^{\pi^{E}}$ is the optimal value function of the expert's MDP utilizing policy $\pi^{E}$.

We propose a method as a baseline for comparison, which we call it most likely trajectory method. Instead of performing robust-IRL, learner can follow another approach. At each time-step $t$ after receiving the observation $o_{t}$ the learner can calculate $Pr(o_{t}|s,a)$ for $\forall s \in S$ and $\forall a \in A$ and then choose the $(s,a)$ for the time-step $t$ with the highest probability. After constructing the trajectory the learner could use the trajectory and learn the reward function. As expected this method is faster than robust-IRL method because it avoids the expectation maximization, however, it is not as accurate as robust-IRL especially under severely noisy conditions.

\subsection{Learning Drone Reconnaissance Routine}
The first domain is a simulation-only domain in which a robot is tasked with learning the policy of a drone that protects a corridor. In this domain, the learner is hidden from the drone's sight. An important challenge in this domain is that the only observation available for the learner is the sound from drone's propellers. The drone follows a policy from its MDP, however, the robot models the drone as an hMDP. The state of the MDP is the location and orientation of the drone in the corridor. The drone has 3 actions, going forward, turn around, hover. drone's transition function modeled as executing the intended action with the probability of 0.9 percent and the remaining probability is uniformly distributed between two remaining actions. Moreover, drone's reward function modeled as a linear combination of following binary features. 

\begin{itemize}
  \item Moved forward: it returns 1 if the drone moves forward, otherwise 0.
  \item Turned around: it return 1 if the drone make a U-turn at state $s$, otherwise 0.
\end{itemize}

\begin{figure}[t]
\centering
\includegraphics[width=1.0\linewidth]{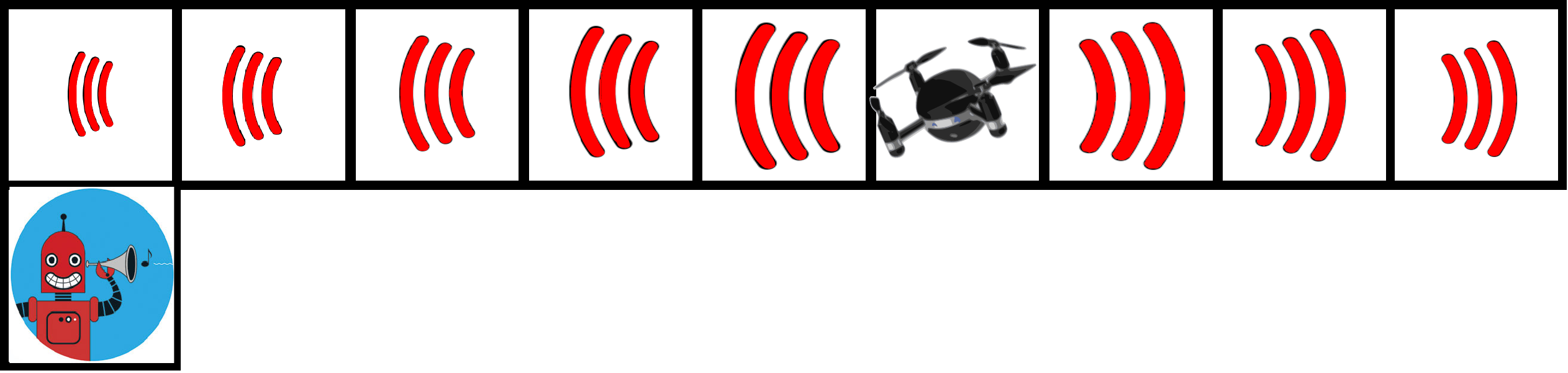}
\caption{\small The robot disguises in the lower left corner and listen to sound from drone's propellers and learns its routine.}
\label{fig:p1}
\end{figure}

The observations in the hMDP are the parameters of the function $f(t)$ described in the Sec.~\ref{sec:obs}. Since in this domain, the learner only receives the expert's movement sound the observation model is solely constructed based on the sound intensity. Figure~\ref{fig:p1} is the illustration of the \textit{learning drone reconnaissance routine} problem domain. 

\begin{figure}[h]
\centering
\includegraphics[width=1.0\linewidth]{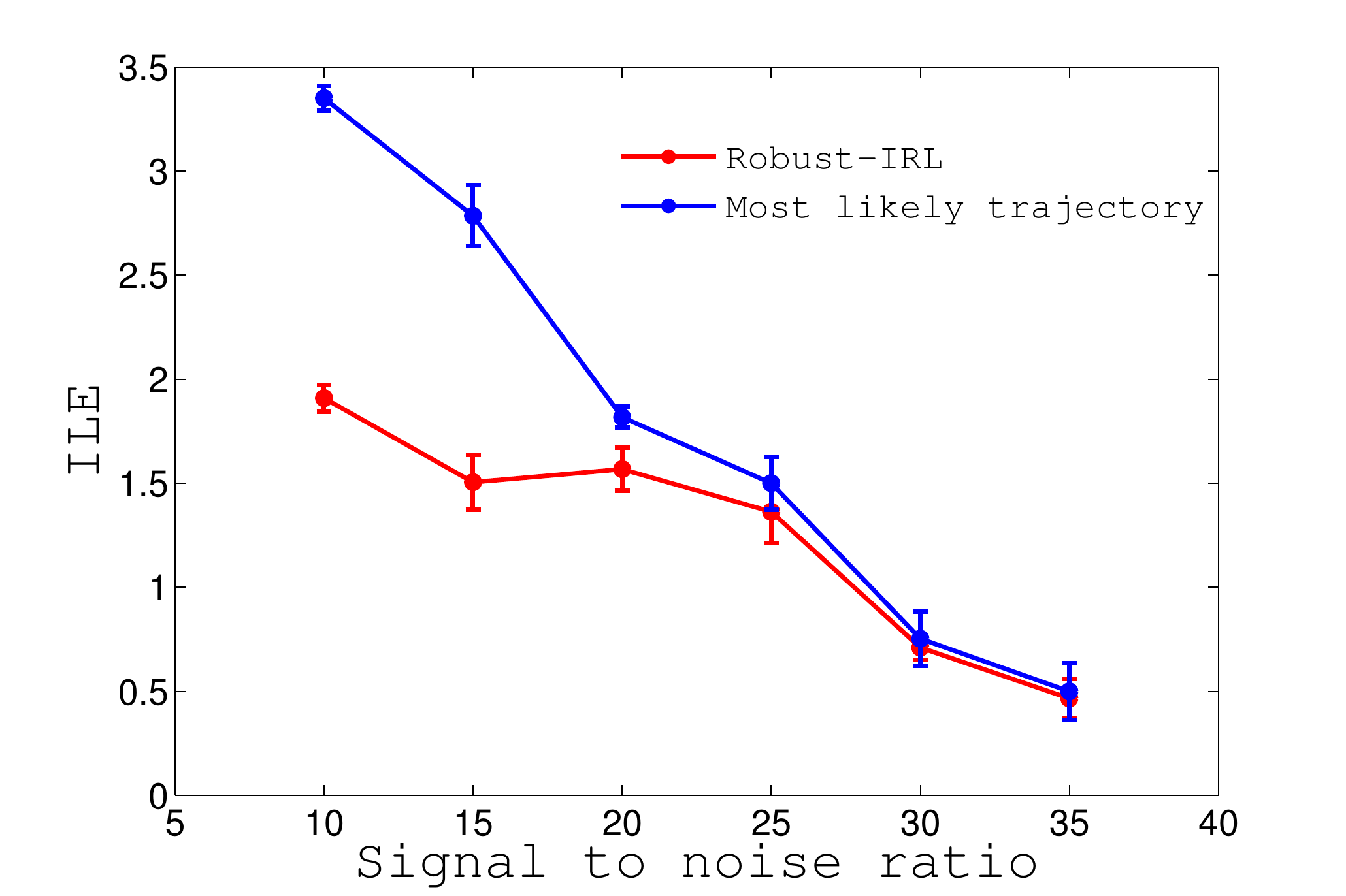}
\caption{\small Performance evaluation of two methods for the learning drone reconnaissance routine simulation. ILE performance for Robust-IRL method and most likely trajectory method. The horizontal axis shows the amount of noise added to the observation. As expected Robust-IRL method has a better performance compare to the other method as the level of noise increases.}
\label{fig:p1-2}
\end{figure}

\subsection{Penetrating a Patrol}
This domain is used for both simulation and physical robot experiment. In this domain, a Turtlebot (intruder) is tasked with learning the policy of another Turtlebot (patroller) that is patrolling an area. The intruder must learn the patroller's policy and reach to the goal state without being seen by the patroller. Both robots in our simulation and physical experiments are Turtlebots equipped with Microsoft XBox 360 Kinect, which provides a camera and a ranging sensor. Moreover, the learner equipped with a microphone that enables it to listen to expert's movement sound. In this domain, the MDP's state is consists of the patroller's location and orientation and the actions are, turning 90 degrees, moving forward, no operation. Patroller's reward function modeled as a linear combination of following binary features. 
\begin{itemize}
  \item Moved forward: it returns 1 if the patroller moves forward, otherwise 0.
  \item Turned around: it return 1 if the patroller makes a U-turn at state $s$, otherwise 0.
\end{itemize}

\begin{figure}
\centering
\includegraphics[width=85mm]{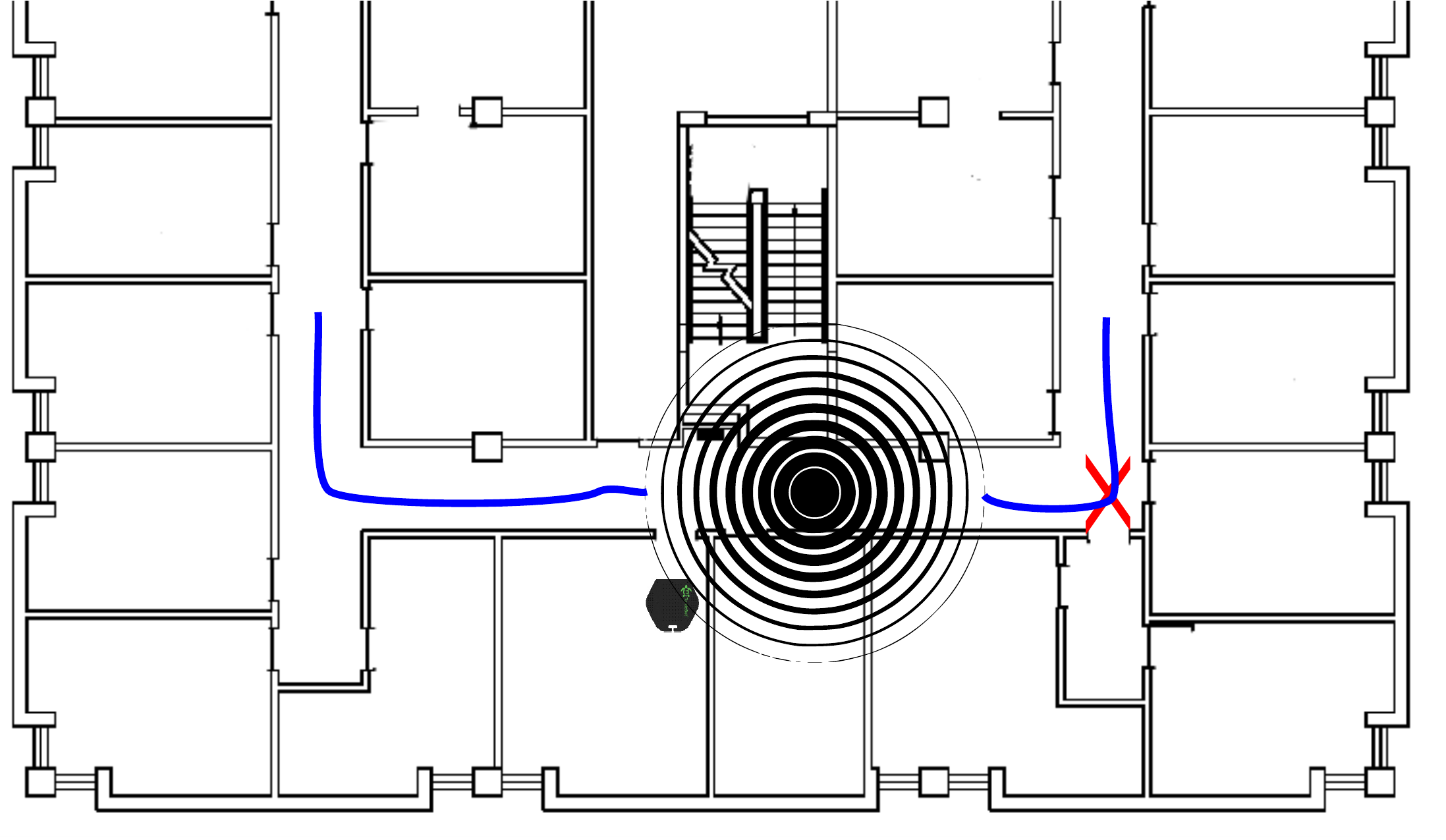}
\caption{\small The penetrator is hidden in the room and the patroller protecting the goal, indicated by X. The penetrator must learn how the patroller moves in the hallway then reach the goal without being seen by the patroller. The blue line indicates the expert's patrolling path, and the concentric circles indicate the magnitude of the sound intensity generated from expert's movement.}
\label{fig:p2}
\end{figure}

In this domain, the observation for the hMDP is consists of sensory data from the Kinect's range finder and sound intensity from the microphone. When the expert is in sight the learner considers the range finder data as observation and when the expert is not in sight the learner considers sound intensities as observations. Moreover, we consider the case where we disregard the sound from the expert and only use the visual information. In this observation model when the expert is not in sight, we consider a uniform distribution over possible state and actions. As expected, utilizing the information from the microphone helps the learner.  Figure~\ref{fig:p2} illustrates \textit{penetrating a patrol} problem domain.

\begin{figure}[h]
\centering
\includegraphics[width=1.0\linewidth]{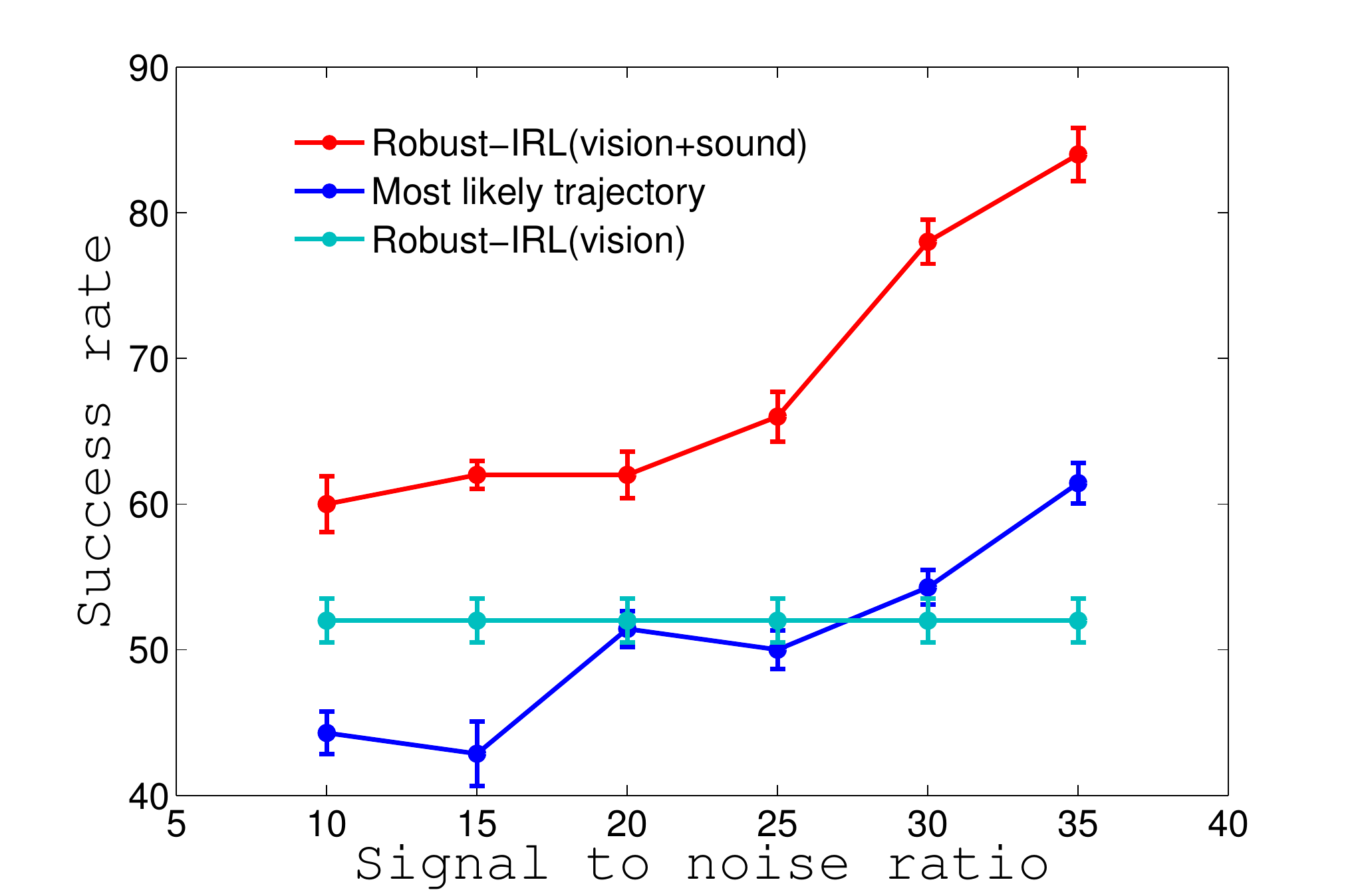}
\caption{\small Performance evaluation of two methods on the Penetrating a patrol domain. This comparison is based on the successful runs. The robust-IRL method is evaluated using two different observation models. The horizontal axis shows the amount of noise added to the sound. As expected robust-IRL method has a better performance compare to the other method as the level of noise increases.}
\label{fig:sr}
\end{figure}

\begin{figure}[!ht]
\centering
\includegraphics[width=85mm]{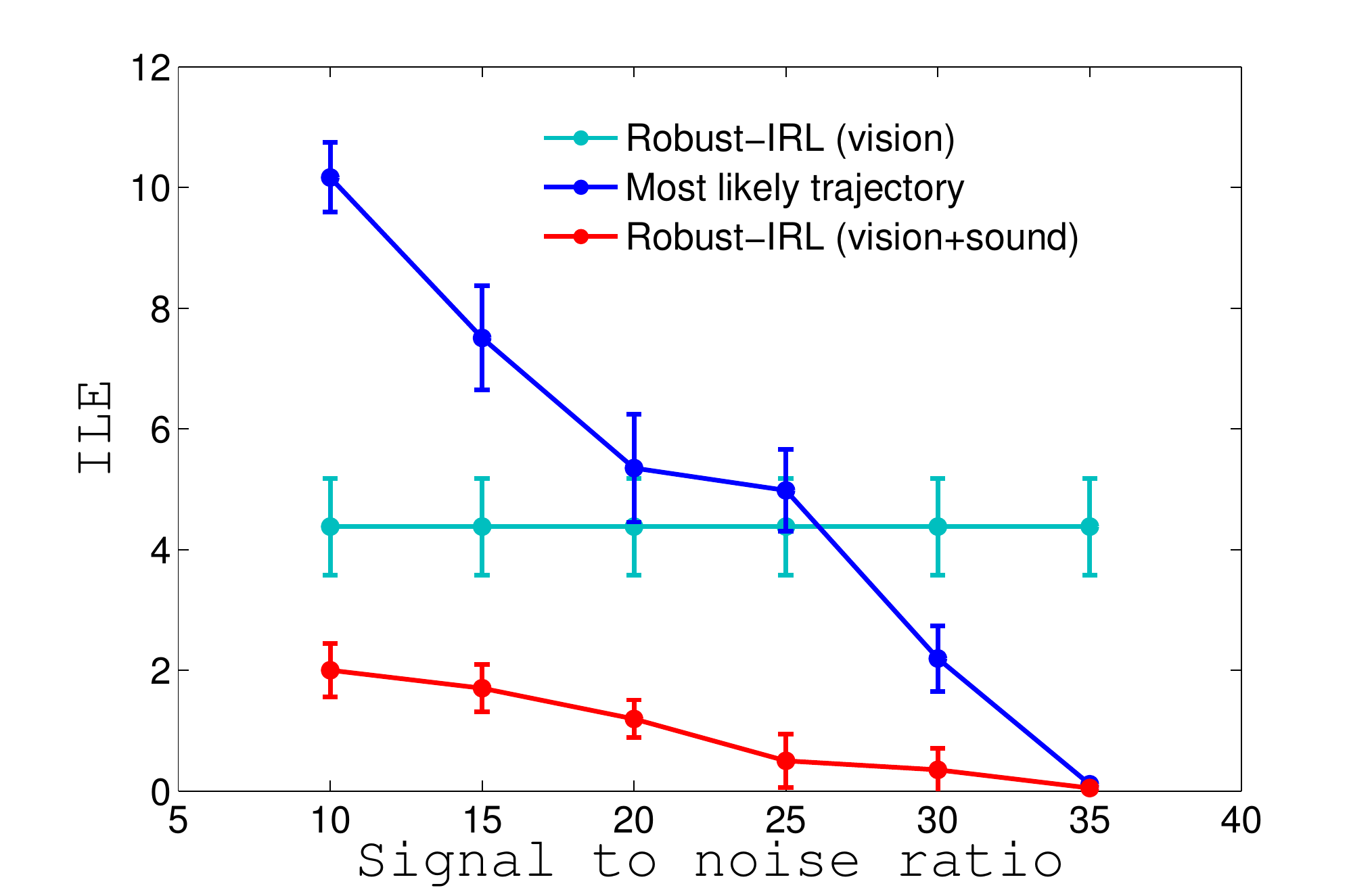}
\caption{\small Performance evaluation of two methods on the penetrating a patrol simulation. ILE performance for Robust-IRL method and using most likely trajectory method. The horizontal axis shows the amount of noise added to the observation. As expected Robust-IRL method has a better performance compare to the other method as the level of noise increases.}
\label{fig:p2-2}
\end{figure}
\begin{figure}[!ht]
\centering
\includegraphics[width=85mm]{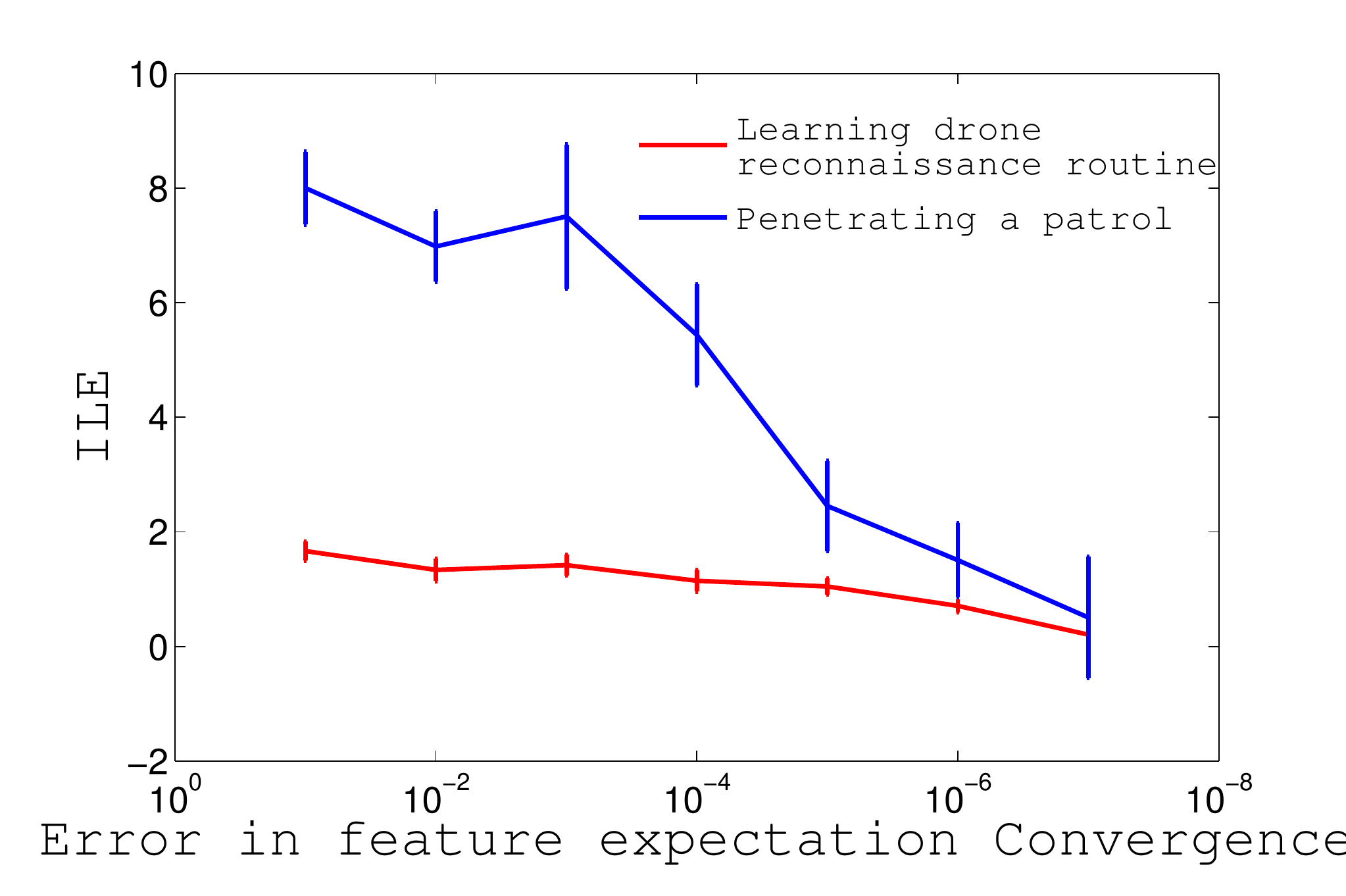}
\caption{\small Performance evaluation of robust-IRL in two domains. The horizontal axis shows the convergence threshold in calculating the feature expectation in the E-step. The robust-IRL method is evaluated using two different observation models. Lower numbers mean a tighter convergence condition. Lower ILE means higher accuracy.}
\label{fig:p12-1}
\end{figure}

\subsubsection{Physical runs}

In addition to simulation for penetrating a patrol domain, we evaluate the performance of robust-IRL with physical robots. Table~\ref{tb:pr} shows the obtained results from 10 physical runs for each method. Due to the limited battery life of Turtlebots, we limited the experiment time to 30 minutes for each run. It means we give the total time of 30 minutes to the learner to observe and learn. If by 30 minutes the learner could reach the goal without being seen by the patroller, we count that as a successful run. All other cases we counted as unsuccessful runs.

For physical runs, we used a random attack approach as an extra baseline. In random attack the learner wait for a random amount of time then it attacks.
\begin{table}[!ht]
\centering
\caption{Results from physical runs}
\begin{tabular}{|l|c|c|} \hline
Method & \begin{tabular}{@{}c@{}}Successful \\ runs\end{tabular} & \begin{tabular}{@{}c@{}}Unsuccessful \\ runs\end{tabular}\\ \hline
Robust-IRL&7&3\\ \hline
Most likely trajectory &4&6\\ \hline
Random Attack &1&9\\ \hline
\end{tabular}\label{tb:pr}
\end{table}

A video of one of our physical runs is available at:\\ \href{http://goo.gl/2DA0lz}{http://goo.gl/2DA0lz}

\section{Related work}\label{sec:rw}

Ng and Russell~\cite{ng2000algorithms} introduced the idea of inverse reinforcement learning as learning the reward function of an expert, modeled as MDP. Later Obermayer and Muckler~\cite{obermayer1965inverse} utilized inverse optimal control to model the experts using other frameworks other than MDP, and Ratliff \textit{at el.}~\cite{ratliff2006maximum} modeled the reward function as a linear combination of features.

Since the introduction of inverse reinforcement learning researchers tries to generalize it by relaxing some of the core assumptions of the IRL problem. Choi and Kim~\cite{choi2011inverse} try to relax the assumption of the full observability of the environment for the expert. However, there has been no in-depth research on the cases, where the observation of the learner is obscured to some extent till Bogert and Doshi~\cite{Bogert:2014:MIR:2615731.2615762} extended the Max-Ent IRL framework to suit multi-agent settings while they allow for occlusion in learner's observation. However, they assumed when the learner receives an observation, it is completely noise-free.

In this paper, we relax the key assumption that the learner's observation is noise free. This has significant implications for making IRL more robust to noise. Bogert and Doshi~\cite{Bogert:2016:EIR:2936924.2937076} also investigates maximum entropy IRL under the presence of latent variable. However, the key difference exists in the assumption of the presence of noise in the observation.

Moreover, Kitani \textit{at el.}~\cite{kitani2012activity} investigate IRL problem under noisy observation. However, key differences exist in the method and assumption that the observation model is dependent on both state and actions. As explained before incorporating actions into observation model introduces challenges that need to be addressed. In comparison, our method is more general and considers cases where we cannot exclude actions from the observation model.

\section{Conclusion}\label{sec:con}

In the context of real world robotics problems presence of noise in observation is usually unavoidable. Our method proposes a mathematical framework to deal with noise. We propose an observation model and try to recover a distribution over trajectories given observation. The experiments show promising results. They indicate that useful policies can be learned by the learner robot even if the amount of noise is considerable.

Also, we show that the learner robot can integrate information from various sensors with different level of accuracy through robust-IRL.

%
%APPENDICES are optional
%\balancecolumns
\appendix
Following is the proof for the Theorem~\ref{th-f(t)}.
\begin{proof}
Suppose that the expert moves from one point to another point.
If:\\
First point coordinate = $(x_0,y_0)$\\
Second point coordinate = $(x,y)$\\
Velocity along the x axis = $v_x$\\
Velocity along the y axis = $v_y$\\
Time at the first point = $t_0$\\
Time at the second point = $t$\\
Then
\begin{center}
$x = v_x(t-t_0)+x_0=v_xt+(x_0-v_xt_0)$
$y = v_y(t-t_0)+y_0=v_yt+(y_0-v_yt_0)$
$r^2=(x-x_0)^2+(y-y_0)^2$
$r^2=(v_xt+(x_0-v_xt_0)-x_0)^2+(v_yt+(y_0-v_yt_0)-y_0)^2$
$r^2=(v_xt-v_xt_0)^2+(v_yt-v_yt_0)^2$
$r^2=v_x^2t^2+v_x^2t_0^2-2v_x^2t_0t+v_y^2t^2+v_y^2t_0^2-2v_y^2t_0t$
$r^2=(v_x^2+v_y^2)t^2+(-2v_x^2t_0-2v_y^2t_0)t+(v_x^2t_0^2+v_y^2t_0^2)$
$r^2=at^2+bt+c$
where:
$a=v_x^2+v_y^2$
$b=-2v_x^2t_0-2v_y^2t_0$
$c=v_x^2t_0^2+v_y^2t_0^2$
$I = \frac{k}{r^2}$
$I = \frac{k}{at^2 + bt + c}$
\end{center} 
\end{proof}

%\section*{References}
\bibliographystyle{abbrv}
\bibliography{sdAAMAS17}
% Generated by bibtex from your ~.bib file.  Run latex,
% then bibtex, then latex twice (to resolve references)
% to create the ~.bbl file.  Insert that ~.bbl file into
% the .tex source file and comment out
% the command \texttt{{\char'134}thebibliography}.
% This next section command marks the start of
% Appendix B, and does not continue the present hierarchy

\end{document}